\documentclass[11pt,a4paper,twocolumn]{article}
\usepackage{latexsym}
\usepackage[T1]{fontenc}

\usepackage{arxiv}
\usepackage{times}  % DO NOT CHANGE THIS
\usepackage{helvet} % DO NOT CHANGE THIS
\usepackage{courier}  % DO NOT CHANGE THIS
\usepackage[hyphens]{url}  % DO NOT CHANGE THIS
\usepackage{graphicx} % DO NOT CHANGE THIS
\usepackage{bm}
\usepackage{xcolor}
\usepackage{graphicx}  % DO NOT CHANGE THIS
\usepackage{multirow}
\usepackage{multicol}
\usepackage{array}
\usepackage{amsfonts}
\usepackage{algpseudocode}
\usepackage{booktabs}
\usepackage{amsmath}
\usepackage{hyperref}

\definecolor{darkblue}{rgb}{0, 0, 0.5}

\hypersetup{
    colorlinks=true,
    linkcolor=darkblue,
    filecolor=darkblue,      
    urlcolor=darkblue,
    citecolor=darkblue,
}

\def\argmax{\mathop{\rm argmax}}
\def\argmin{\mathop{\rm argmin}}

\title{\textbf{Learning to Predict Explainable Plots \\ for Neural Story Generation}}

\author{\textbf{Gang Chen}$^\dagger$, \textbf{Yang Liu}$^{\dagger\ddagger}$, \textbf{Huanbo Luan}$^\dagger$, \textbf{Meng Zhang}$^\#$, \\ \textbf{Qun Liu}$^\#$, \textbf{and Maosong Sun}$^{\dagger}$ \\
 \\	
	$^\dagger$Institute for Artificial Intelligence \\
    State Key Laboratory of Intelligent Technology and Systems \\
    Department of Computer Science and Technology, Tsinghua University, Beijing, China \\
    $^\ddagger$Beijing National Research Center for Information Science and Technology \\
    $^\#$Noah's Ark Lab Paris, Huawei Technologies Ltd
}

\date{}

\begin{document}

\maketitle

\begin{abstract}
Story generation is an important natural language processing task that aims to generate coherent stories automatically. While the use of neural networks has proven effective in improving story generation, how to learn to generate an explainable high-level plot still remains a major challenge. In this work, we propose a latent variable model for neural story generation. The model treats an outline, which is a natural language sentence explainable to humans, as a latent variable to represent a high-level plot that bridges the input and output. We adopt an external summarization model to guide the latent variable model to learn how to generate outlines from training data. Experiments show that our approach achieves significant improvements over state-of-the-art methods in both automatic and human evaluations.
\end{abstract}

\section{Introduction}

\begin{figure}[!t]
\centering
\includegraphics[width=0.45\textwidth]{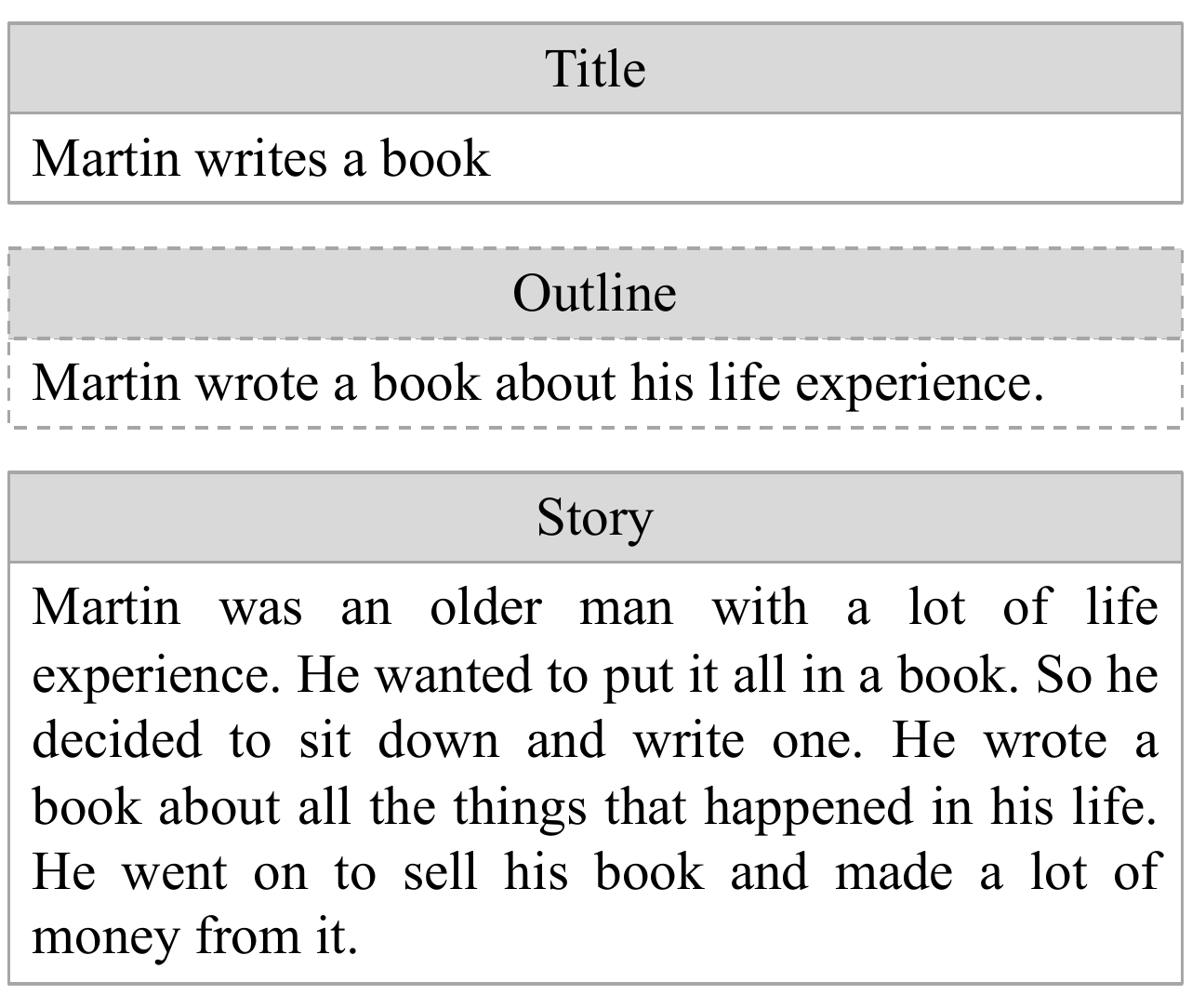}
\caption{Neural story generation aims to generate a story given a title. The training data only contains pairs of titles and stories. Our work treats the outline that bridges the title and the story as a latent variable. Note that our work differs from previous work in that the outlines are represented as natural language sentences explainable to humans rather than real-valued vectors.}
\label{fig:example}
\end{figure}

% story generation
Story generation, which aims to generate coherent stories automatically, is an important task in natural language processing (NLP). While early efforts have focused on symbolic and logical planning \cite{mcintyre:acl09,Riedl:10,Martens:14} and case-based reasoning \cite{Turner:94,Gervas:05}, neural story generation (NSG) that leverages neural networks to generate stories has received increasing attention recently \cite{Roemmele:16,yao:aaai18,Martin:18,Fan:18,Xu:18,subramanian:nips18,moryossef:naacl2019,goldfarbtarrant:naacl19}.

% information gap
Despite its rapid development, NSG still suffers from an ``{\em information gap}'' problem: it is hard to generate a long output given a short input. Figure~\ref{fig:example} shows an example. Given a title ``Martin writes a book'', NSG needs to generate a coherent and fluent story accordingly. As the story is much longer than the title, it is challenging for NSG to ensure that the story is centered on the title and remains thematically consistent. This problem has been considered as one of the key challenges in story generation \cite{Wiseman:17}. 

% related work
Although a number of authors have endeavored to address this problem by introducing high-level plots (i.e., outlines) to bridge titles and stories \cite{Fan:18,Xu:18,Drissi:18,yao:aaai18} (see Section \ref{sec:hnsg} for details), there still remain two unsolved problems. First, it is hard to learn the latent outlines. As the training data only contains titles and stories, outlines are unobserved. While it is natural to use a latent variable model \cite{Kim:18} and treat outline as a latent variable, representing outlines as real-valued vectors makes it difficult for humans to interpret how outlines connect titles and stories. On the contrary, representing outlines as natural language sentences improves interpretability but faces the exponential search space of outlines.

Second, the lack of exposure to degenerate outlines impairs the generation performance. \citet{Drissi:18} indicate that their approach is incapable of dealing with poorly generated outlines because the generation model is only trained on good outlines in a separate way. When the generated outline is erroneous, the generation model is unlikely to be able to recover from the mistake.

% our work
In this work, we propose a method for learning to predict explainable high-level plots for NSG to alleviate the two aforementioned problems. In our model, the outline unobserved on the training data is treated as a latent variable. Given a training set that contains only pairs of titles and stories, we adopt an external summarization model to instruct the latent variable model to learn how to generate outlines. We also use a joint training objective to avoid explicit enumeration of exponentially many outlines and make the model exposed to predicted high-level plots (see Eq.~(\ref{eq:train_obj})). Taking advantage of the generated outline to fill the information gap between the title and story, our approach can produce better stories compared with existing methods. Experiments on two widely used datasets show that our approach achieves significant improvements over existing methods in both automatic and human evaluations.

\section{Related Work}
Our work is closely related to two lines of research: hierarchical neural story generation and variational neural networks.

\subsection{Hierarchical Neural Story Generation} \label{sec:hnsg}

Neural story generation has received increasing attention in recent years \cite{Roemmele:16,Jain:17,Harrison:aaai17,Wang:acl18,Martin:18,Xu:18,Fan:18}. Unlike other sequence-to-sequence learning tasks such as machine translation \cite{Bahdanau:iclr15,Vaswani:17} and dialogue generation \cite{Serban:aaai16}, in which the input and output sequences are often of similar lengths, one major difficulty in neural story generation is that the output sequence is much longer than the input sequence. As a result, hierarchical models for neural story generation have been intensively studied recently \cite{Xu:18,Fan:18,Drissi:18,subramanian:nips18,moryossef:naacl2019,goldfarbtarrant:naacl19,yao:aaai18,fan:acl19}.

Our work is closest to the hierarchical model proposed by \citet{Drissi:18}, which also uses an outline as a high-level plot to reduce the information gap between input and output sequences. The major difference is that we treat outline generation as a learning task while \citet{Drissi:18} use an outline produced by an off-the-shelf text summarizer. Our experiments reveal that learning to generate outlines significantly improves over directly using the output of an off-the-shelf text summarizer. 

This work also resembles the method proposed by \citet{Xu:18} because both skeletons and outlines are learned from data automatically. The difference is that we use outlines to bridge titles and stories while \citet{Xu:18} focus on using skeletons to better capture the dependencies between sentences in stories.

Our approach is also related to the method proposed by \citet{yao:aaai18}, which generates a storyline from title to help generate the story. The main difference is that the outlines in our method consists of natural sentences while the model proposed by \citet{yao:aaai18} learns to generate a sequence of keywords as the storyline.

\subsection{Variational Neural Networks}

Variational neural networks \cite{Graves:11,Kingma:13,Mnih:14} have been widely used for many NLP tasks in recent years. \citet{Bowman:16} propose a generative model that uses a latent variable to learn the global feature for sentence representation. \citet{VNMT:acl16} design a variational encoder-decoder framework that leverages a latent variable to model the underlying semantics of source sentences. \citet{calixto:acl18} incorporate image features  to neural machine translation using latent variables.

Our work differs from prior studies on variational neural networks in that the latent variable is represented as discrete symbols rather than continuous vectors. This makes latent variables in our model interpretable to humans. It is easy for us to observe how the model learns the latent variables during training (see Table~\ref{tab:analysis_outline}). This is beneficial for analyzing and understanding neural network based NSG models.

\begin{figure}[!t]
\centering
\includegraphics[width=0.45\textwidth]{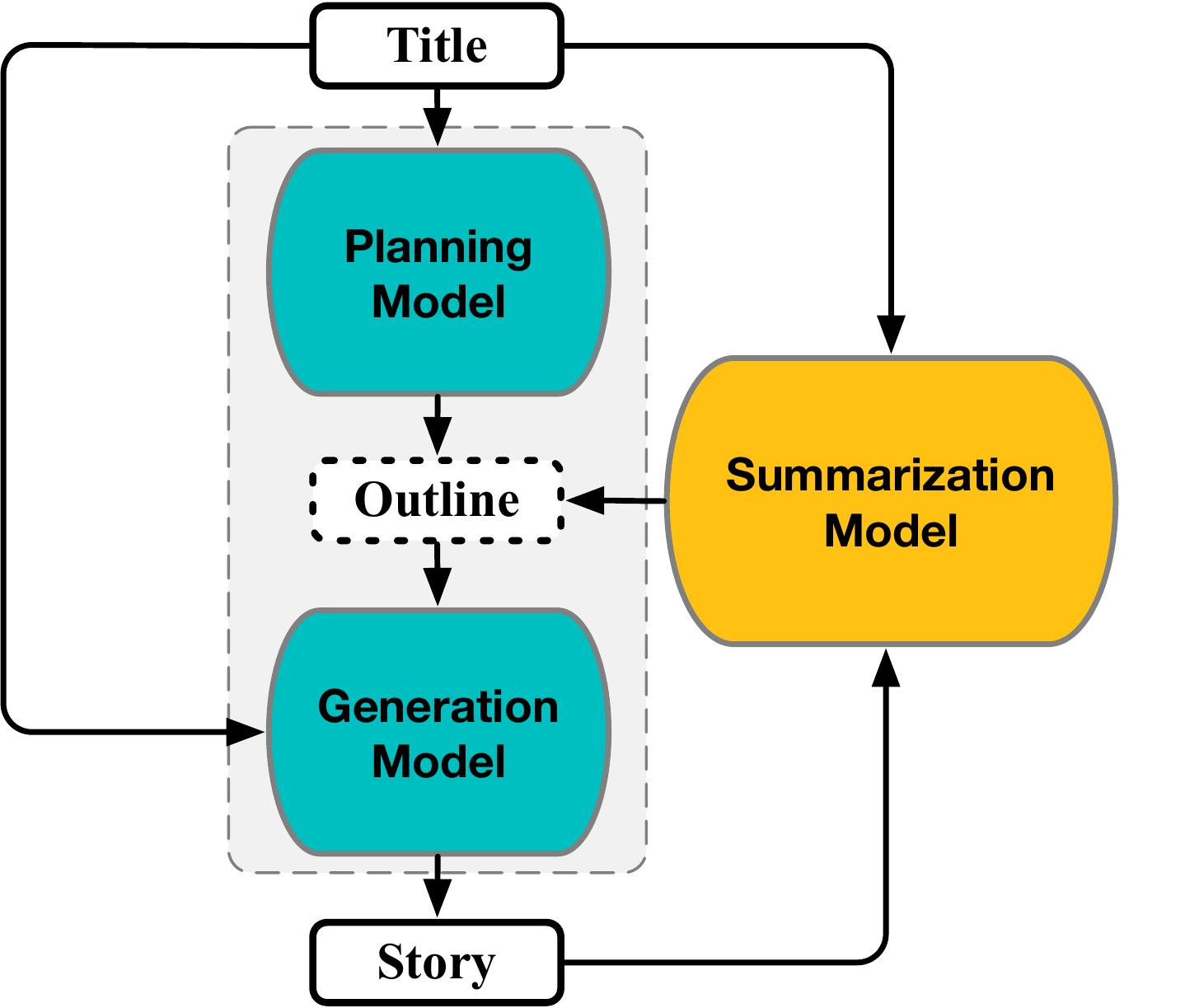}
\caption{An overview of our latent variable model for NSG. Note that the summarization model is only used in the training phase while the planning model and the generation model are involved in both the training and inference phases.} \label{fig:model}
\end{figure}

\section{Approach} \label{sec:approach}

Figure~\ref{fig:model} shows an overview of our model. Given a title, we use a {\em planning model} to generate an outline, from which a {\em generation model} runs to produce the story (Section~\ref{sec:model}). As outlines are unobserved on the training data, we use a {\em summarization model} to ``teach'' the planning model how to generate outlines. We use the outlines produced by off-the-shelf text summarizers to pre-train the three models, which are then jointly trained on the training data (Section~\ref{sec:training}). Finally, the learned planning and generation models can be used to generate new stories for unseen titles (Section~\ref{sec:generation}).

\subsection{Modeling} \label{sec:model}

\subsubsection*{The Latent Variable Model}

Let $\mathbf{x} = x_1 \dots x_I$ be a title that contains $I$ words, which is the input of neural story generation. We use $\mathbf{y} = y_1 \dots y_J$ to denote a story that contains $J$ words, which is the output of neural story generation. As the stories in most widely used datasets such as ROCStories \cite{Mostafazadeh:16} and VIST \cite{Huang:16} are usually very short as shown in Figure~\ref{fig:example}, we assume there is only a one-sentence outline $\mathbf{z}=z_1 \dots z_K$ that contains $K$ words for the entire story.

The latent variable model for neural story generation is given by

\begin{eqnarray}
P(\mathbf{y}|\mathbf{x}; \bm{\theta}, \bm{\gamma}) = \sum_{\mathbf{z}} \underbrace{P(\mathbf{z}|\mathbf{x}; \bm{\theta})}_{\mathrm{planning}} \underbrace{P(\mathbf{y}|\mathbf{x}, \mathbf{z}; \bm{\gamma})}_{\mathrm{generation}}, \label{eq:latent_model}
\end{eqnarray}
where $P(\mathbf{z}|\mathbf{x}; \bm{\theta})$ is the {\em planning model} parameterized by $\bm{\theta}$ and $P(\mathbf{y|\mathbf{x}, \mathbf{z}}; \bm{\gamma})$ is the {\em generation model} parameterized by $\bm{\gamma}$.

\subsubsection*{The Planning Model}

The planning model is used to generate an outline given a title:

\begin{eqnarray}
P(\mathbf{z} | \mathbf{x}; \bm{\theta}) = \prod_{k=1}^{K} P(z_k | \mathbf{x}, \mathbf{z}_{<k}; \bm{\theta}),
\end{eqnarray}
where $\mathbf{z}_{<k} = z_1 \dots z_{k-1}$ is a partial outline. 

We use Transformer \cite{Vaswani:17}, which is the state-of-the-art sequence-to-sequence model for neural machine translation, to build the planning model. The title is the input of Transformer and the outline is the output.

\begin{figure}[!t]
\centering
\includegraphics[width=0.48\textwidth]{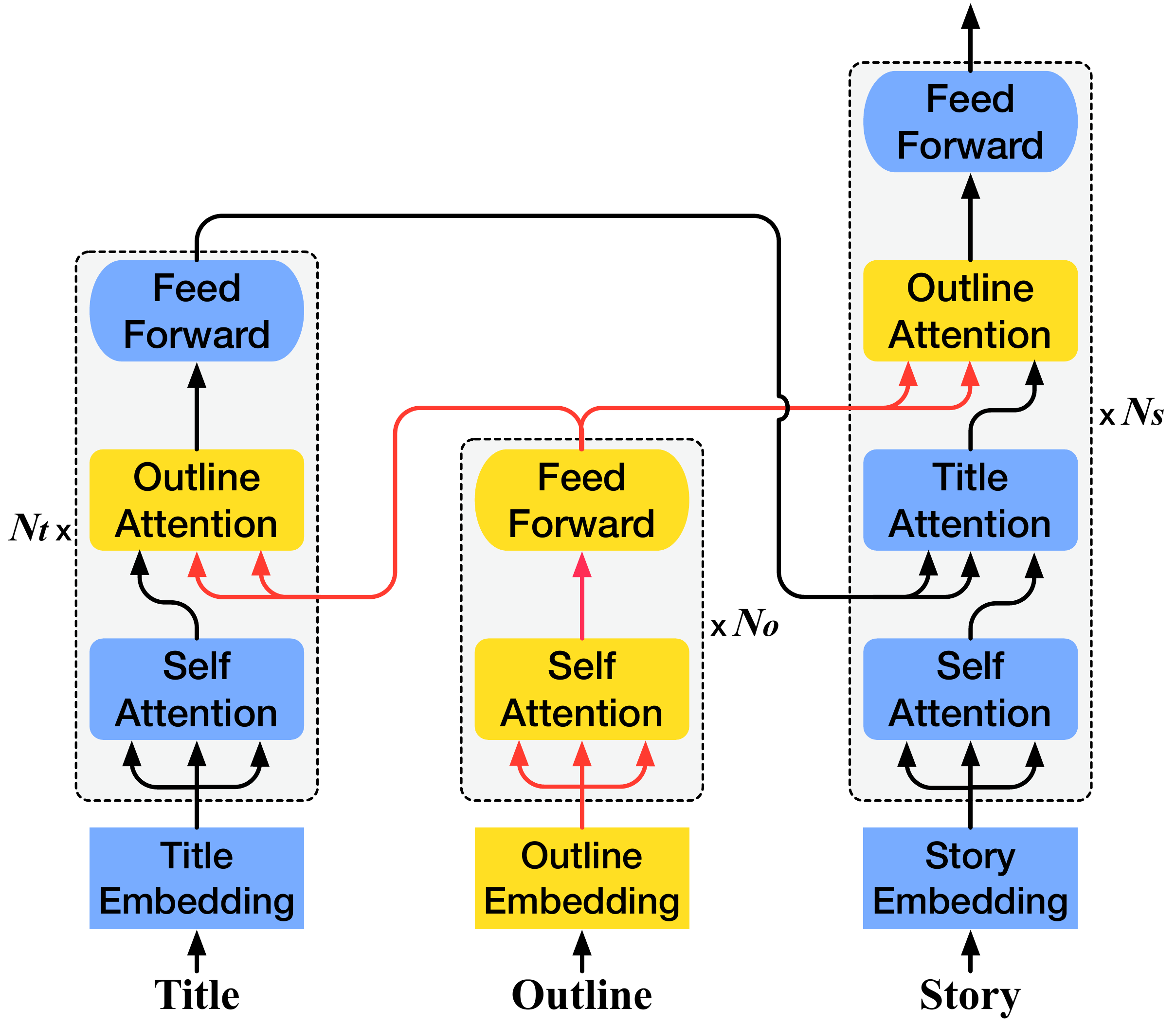}
\caption{The architecture of the generation model.} \label{fig:generation}
\end{figure} 

\subsubsection*{The Generation Model}

The generation model is used to generate a story given a title and an outline:

\begin{eqnarray}
P(\mathbf{y}|\mathbf{x}, \mathbf{z}; \bm{\gamma}) = \prod_{j=1}^{J}P(y_j | \mathbf{x}, \mathbf{z}, \mathbf{y}_{<j}; \bm{\gamma}),
\end{eqnarray}
where $\mathbf{y}_{<j} = y_1 \dots y_{j-1}$ is a partial story.

The architecture of the generation model is similar to that  of the planning model except that the input of the generation model has two sources, namely the title $\mathbf{x}$ and the outline $\mathbf{z}$. We expect that the generation model is able to take advantage of the information contained in the outline to guide the story generation. 
Inspired by \cite{Zhang:18}, which proposes a strategy to utilize the document-level context to improve sentence-level neural machine translation, we extend the vanilla sequence-to-sequence framework to a multi-source architecture that can make use of both titles and outlines.  Figure~\ref{fig:generation} shows the architecture of the generation model. As indicated by the red arrows, the outline is not only used to learn a better representation of the title, but also participates in the generation of the story. \citet{Zhang:18} show that such an architecture is capable of effectively exploiting all sources of input. Note that the model reduces to the standard title-to-story architecture if the yellow components related to outlines are removed.

\subsection{Training} \label{sec:training}

\subsubsection*{Training Objective}

It is challenging to train the latent variable model in Eq.~(\ref{eq:latent_model}) because there are exponentially many possible outlines for a given title. As the full enumeration of all outlines is impossible and approximations have to be made, it is necessary to leverage useful external information to facilitate the learning of latent variables. % We use {\em posterior regularization}  %Similar with variational inference \cite{Kingma:13,Rezende:14},
The training objective of our model is given by

\begin{eqnarray}
&& J(\mathbf{x}, \mathbf{y}, \bm{\theta}, \bm{\gamma}, \bm{\phi}) \nonumber \\
&=& \mathbb{E}_{\mathbf{z}|\mathbf{x}; \bm{\theta}}\Big[ \log P(\mathbf{y} | \mathbf{x}, \mathbf{z}; \bm{\gamma}) \Big] \nonumber \\
&& - \mathrm{KL}\Big( \underbrace{P(\mathbf{z}|\mathbf{x}, \mathbf{y}; \bm{\phi})}_{\mathrm{summarization}} \big|\big| P(\mathbf{z}|\mathbf{x}; \bm{\theta})\Big), \label{eq:train_obj}
\end{eqnarray}
where $P(\mathbf{z}|\mathbf{x}, \mathbf{y}; \bm{\phi})$ is the {\em summarization model} that generates the outline $\mathbf{z}$ based on the title $\mathbf{x}$ and the story $\mathbf{y}$. 

The first term of the right-hand side of Eq.~(\ref{eq:train_obj}) is the expectation of the generation probability, which aims to make the model generate better stories using the titles and the outlines produced by the planning model. The second term is the Kullback-Leibler divergence between the summarization model and the planning model, which suggests that the summarization model is used to ``teach'' the planning model how to generate outlines. This term is also seen as a posterior regularization \cite{ganchev:2010} where the summarization model act as the posterior that gives supervision for the planning model. 

Note that Eq.~(\ref{eq:train_obj}) is similar but different from the lower bound widely used in variational inference \cite{Kingma:13,Rezende:14}. The expectation in Eq.~(\ref{eq:train_obj}) would be defined with respect to $P(\mathbf{z}|\mathbf{x},\mathbf{y};\bm{\phi})$ instead of $P(\mathbf{z}|\mathbf{x}; \bm{\theta})$ in variational inference. We did not directly use variational inference because it is the planning model rather than the summarization model that is used during inference (see Section \ref{sec:generation}). Therefore, we are interested in the expectation with respect to the planning model. Our experiment shows that our approach improves over variational inference on the NSG task (see Table~\ref{tab:train_obj}).\label{fn:vi}

The summarization model can be defined as
\begin{eqnarray}
P(\mathbf{z} | \mathbf{x}, \mathbf{y}; \bm{\phi}) = \prod_{k=1}^{K} P(z_k | \mathbf{x}, \mathbf{y}, \mathbf{z}_{<k}; \bm{\phi}).
\end{eqnarray}
We use an architecture similar to that in Figure~\ref{fig:generation}. The major difference is that the title and the story serve as the input and the outline is the output.

\subsubsection*{Pre-training}

Pre-training has proven to be important for providing a good starting point for parameter estimation in natural language processing \cite{Devlin:18}. We follow \citet{Drissi:18} to use an off-the-shelf text summarizer to generate raw outlines, which are used to pre-train our models.

Let $\{ \langle \mathbf{x}^{(s)}, \mathbf{y}^{(s)} \rangle \}_{s=1}^{S}$ be the training data that contains $S$ pairs of titles and stories. We can use an off-the-shelf text summarizer to generate a set of outlines $\{ \tilde{\mathbf{z}}^{(s)} \}_{s=1}^{S}$ given all the stories. We use these outlines to pre-train the planning, generation, and summarization models by maximum likelihood estimation:

\begin{eqnarray}
\bm{\theta}_0 = \argmax_{\bm{\theta}} \Bigg\{ \sum_{s=1}^{S} \log P(\tilde{\mathbf{z}}^{(s)} | \mathbf{x}^{(s)}; \bm{\theta}) \Bigg\}, \quad \quad \ \\
\bm{\gamma}_0 = \argmax_{\bm{\gamma}} \Bigg\{ \sum_{s=1}^{S} \log P(\mathbf{y}^{(s)} | \mathbf{x}^{(s)}, \tilde{\mathbf{z}}^{(s)}; \bm{\gamma}) \Bigg\}, \\
\bm{\phi}_0 = \argmax_{\bm{\phi}} \Bigg\{ \sum_{s=1}^{S} \log P(\tilde{\mathbf{z}}^{(s)} | \mathbf{x}^{(s)}, \mathbf{y}^{(s)}; \bm{\phi}) \Bigg\}.
\end{eqnarray}

\subsubsection*{Parameter Estimation}

Both of the two terms in the right-hand side of Eq.~(\ref{eq:train_obj}) require summing up all outlines to calculate expectations. Note that KL divergence can also be seen as an expectation. Due to the exponential space of outlines, we use Monte Carlo sampling to approximate the expectations. We draw $M$ samples $\{\mathbf{z}^{(1)},\dots,\mathbf{z}^{(M)}\}$ from the planning model $P(\mathbf{z} | \mathbf{x}; \bm{\theta})$ and draw $N$ samples $\{\mathbf{z}^{(1)}, \dots, \mathbf{z}^{(N)}\}$ from the summarization model $P(\mathbf{z} | \mathbf{x}, \mathbf{y}; \bm{\phi})$. 

As a result, the training objective can be approximated as
\begin{eqnarray}
&& J(\mathbf{x}, \mathbf{y}, \bm{\theta}, \bm{\gamma}, \bm{\phi}) \nonumber \\
&\approx& \frac{1}{M} \sum_{m=1}^{M} \log P(\mathbf{y}|\mathbf{x}, \mathbf{z}^{(m)}; \bm{\gamma}) \nonumber \\
&& - \frac{1}{N} \sum_{n=1}^{N} \log \frac{P(\mathbf{z}^{(n)}|\mathbf{x}, \mathbf{y}; \bm{\phi})}{P(\mathbf{z}^{(n)}|\mathbf{x}; \bm{\theta})}.
\label{Eq:approximate}
\end{eqnarray}

Given a training set $\{ \langle \mathbf{x}^{(s)}, \mathbf{y}^{(s)} \rangle \}_{s=1}^{S}$, the final loss function for training our model can be formulated as
\begin{eqnarray}
L(\bm{\theta}, \bm{\gamma}, \bm{\phi}) = -\sum_{s=1}^{S} J(\mathbf{x}^{(s)}, \mathbf{y}^{(s)}, \bm{\theta}, \bm{\gamma}, \bm{\phi}).
\end{eqnarray}

Initialized with the pre-trained model parameters $\bm{\theta}_0$, $\bm{\gamma}_0$, and $\bm{\phi}_0$, our latent variable model can be optimized as follows:

\begin{eqnarray}
\hat{\bm{\theta}}, \hat{\bm{\gamma}}, \hat{\bm{\phi}} = \argmin_{\bm{\theta}, \bm{\gamma}, \bm{\phi}} \Big\{ L(\bm{\theta}, \bm{\gamma}, \bm{\phi}) \Big\}.
\end{eqnarray}

\subsection{Inference} \label{sec:generation}

The generation process is divided into two steps. Given a new title $\mathbf{x}$, our model first generates an outline using the planning model:

\begin{eqnarray}
\hat{\mathbf{z}} = \argmax_{\mathbf{z}}\Big\{ P(\mathbf{z}|\mathbf{x}; \hat{\bm{\theta}}) \Big\}.
\end{eqnarray}

Then, the generation model is used to generate the story given the title and the outline:

\begin{eqnarray}
\hat{\mathbf{y}} = \argmax_{\mathbf{y}} \Big\{ P(\mathbf{y} | \mathbf{x}, \hat{\mathbf{z}}; \hat{\bm{\gamma}}) \Big\}.
\end{eqnarray}

Note that the summarization model is only involved in training and is not used in inference.

\section{Experiments}
\subsection{Datasets}
We evaluated our approach on two widely used datasets: the ROCStories dataset \cite{Mostafazadeh:16} and the VIST dataset \cite{Huang:16}.

The ROCStories dataset contains 98,161 short stories as the training set and additional 1,871 stories as validation and test sets, respectively. Since only the training set of the ROCStories dataset contains titles, which we need as input, we split the original training set into 8:1:1 for training, validation, and testing. Therefore, the training set we used in the experiments contains 78,529 pairs of titles and stories, the validation set and the test set both contain 9,816 pairs.

The VIST dataset contains pairs of photo sequences and corresponding coherent hand-crafted narratives of events through time. Sentences with the same story ID in the original dataset are concatenated to form a story. Since all the stories in this dataset have no titles, we followed \citet{Xu:18} to use the first sentence of an original story as the title and the remaining sentences as the story. The processed dataset contains 40,155, 4,990, and 5,055 pairs of titles and stories for training, validation, and testing, respectively.

\subsection{Baselines}
We compared our approach with the following NSG methods:

\begin{enumerate}
\item \textproc{Direct} \cite{Roemmele:16}: Directly generating a story given a title using sequence-to-sequence models.

\item \textproc{Skeleton} \cite{Xu:18}: First generating a skeleton and then expanding the skeleton into a complete sentence until the whole story has been generated.

\item \textproc{Hierarchical} \cite{Fan:18}: First generating a prompt and then generating a story based on the prompt.

\item \textproc{Separate} \cite{Drissi:18}: First generating an outline and then generating a story based on the outline using two separate modules.

\item \textproc{Planned} \cite{yao:aaai18}: First generating a storyline and then generating a story based on the title and the storyline.
\end{enumerate}

For \textproc{Direct}, we used Transformer \cite{Vaswani:17} to directly map titles into stories. For \textproc{Skeleton}, we directly used the code \footnote{\url{https://github.com/lancopku/Skeleton-Based-Generation-Model}} released by \citet{Xu:18}. For \textproc{Hierarchical}, we directly used the code \footnote{\url{https://github.com/pytorch/fairseq}} released by \citet{Fan:18}. For \textproc{Separate}, we used two separate Transformers to model the transition from titles to outlines and the transition from outlines to stories, respectively. For \textproc{Planned}, we directly used the code \footnote{\url{https://bitbucket.org/VioletPeng/language-model}} released by \citet{yao:aaai18}, and more specifically, we adopted the static planning scheme. Please refer to \cite{yao:aaai18} for more details.

For convenience, we refer to our approach as \textproc{Latent}. The planning, generation, and summarization models are built on top of the Transformer architecture. The encoders of the planning and generation models are composed of a stack of two identical layers while the encoder of the summarization model comprises a stack of six identical layers. All the decoders of the three models consist of a stack of six identical layers. We set $M$ and $N$ in Eq.~(\ref{Eq:approximate}) to 1, which is good enough in our experiments. Our approach also used the text summarizer proposed by \citet{Nenkova:05} to extract outlines from stories and pre-train our models.

We used beam search to produce stories and set the beam size to 4 in the inference phase. For fair comparisons, we set the hidden size and embedding size to 256, and maximum vocabulary size to 50K for all the baselines and our approach. We set the dimension of the feed-forward network to 1,024 for \textproc{Direct}, \textproc{Separate}, and our method.

\subsection{Evaluation Metrics}
Both automatic and human evaluations are conducted in our experiments.

\begin{table*}[htbp]
    \centering
    \scalebox{1}{
    \begin{tabular}{|l||cccc|cccc|}
    \hline
        \multirow{2}{*}{Model} & \multicolumn{4}{c|}{ROCStories} & \multicolumn{4}{c|}{VIST} \\
        \cline{2-9}
        & {~B-4 $\uparrow$} & {~~MET $\uparrow$} & {~~R-L $\uparrow$} & {~~PPL $\downarrow$} & {~B-4 $\uparrow$} & {~~MET $\uparrow$} & {~~R-L $\uparrow$} & {~~PPL $\downarrow$}  \\
    \hline \hline
    \textproc{Direct} & 3.01 & \,~9.95 & 22.09 & 14.83 & 2.88 & \,~8.27 & 20.59 & 38.72 \\
    \textproc{Skeleton} & 1.45 & \,~7.43 & 19.82 & 29.17 & 0.96 & \,~5.82 & 18.20 & 46.53  \\
    \textproc{Hierarchical} & 2.09 & \,~9.65 & 20.66 & 25.79 & 2.55 & \,~9.34 & 21.10 & 38.05  \\
    \textproc{Separate} & 2.55 & \,~9.44 & 21.47 & 31.30 & 2.75 & \,~8.04 & 20.40 & 58.54   \\
    \textproc{Planned} & 2.87 & \,~9.98 & 22.30 & 28.67 & 2.61 & \,~9.63 & 20.79 & 40.43 \\
    \hline
    \textproc{Latent} & \textbf{3.70} & \textbf{10.61} & \textbf{23.06}  & \textbf{12.33} & \textbf{3.35} & \textbf{10.03} & \textbf{21.76} & \textbf{36.34}  \\
    \hline
    \end{tabular}
    }
    \caption{Automatic evaluation on the ROCStories and VIST datasets. We use ``B-4'', ``MET'', ``R-L'', and ``PPL'' to denote BLEU-4, METEOR, ROUGE-L, and perplexity scores, respectively.}
    \label{tab:auto_results}
\end{table*}

\subsubsection*{Automatic Evaluation Metrics}
Following existing methods \cite{Guan:aaai19,Wang:acl18short}, we use BLEU-4 \cite{Bleu:acl02}, METEOR \cite{meteor:acl05}, ROUGE-L \cite{rouge:04}, and model perplexity as automatic evaluation metrics. BLEU and ROUGE-L encourage exact match between ground-truth text and generated text by $n$-gram overlap, and METEOR also uses WordNet stems and synonyms. Perplexity is designed to evaluate the quality of language models because it reflects how fluently the model can produce the correct next word given the preceding words. Smaller perplexity indicates better performance.

\subsubsection*{Human Evaluation Metrics}
As \citet{Wang:acl18} state that there are no perfect automatic evaluation metrics to quantify the quality of generated stories, it is important to use human evaluation metrics. In our experiments, we used three human evaluation metrics: {\em fluency}, {\em relevance}, and {\em coherence}. Fluency measures whether each sentence in the story is grammatically correct. Relevance measures whether the story is related to the title. Coherence evaluates whether the story is coherent. Each metric score ranges from 1 to 10. Higher score indicates better performance.

\begin{figure}[t]
    \centering
    \includegraphics[width=0.48\textwidth]{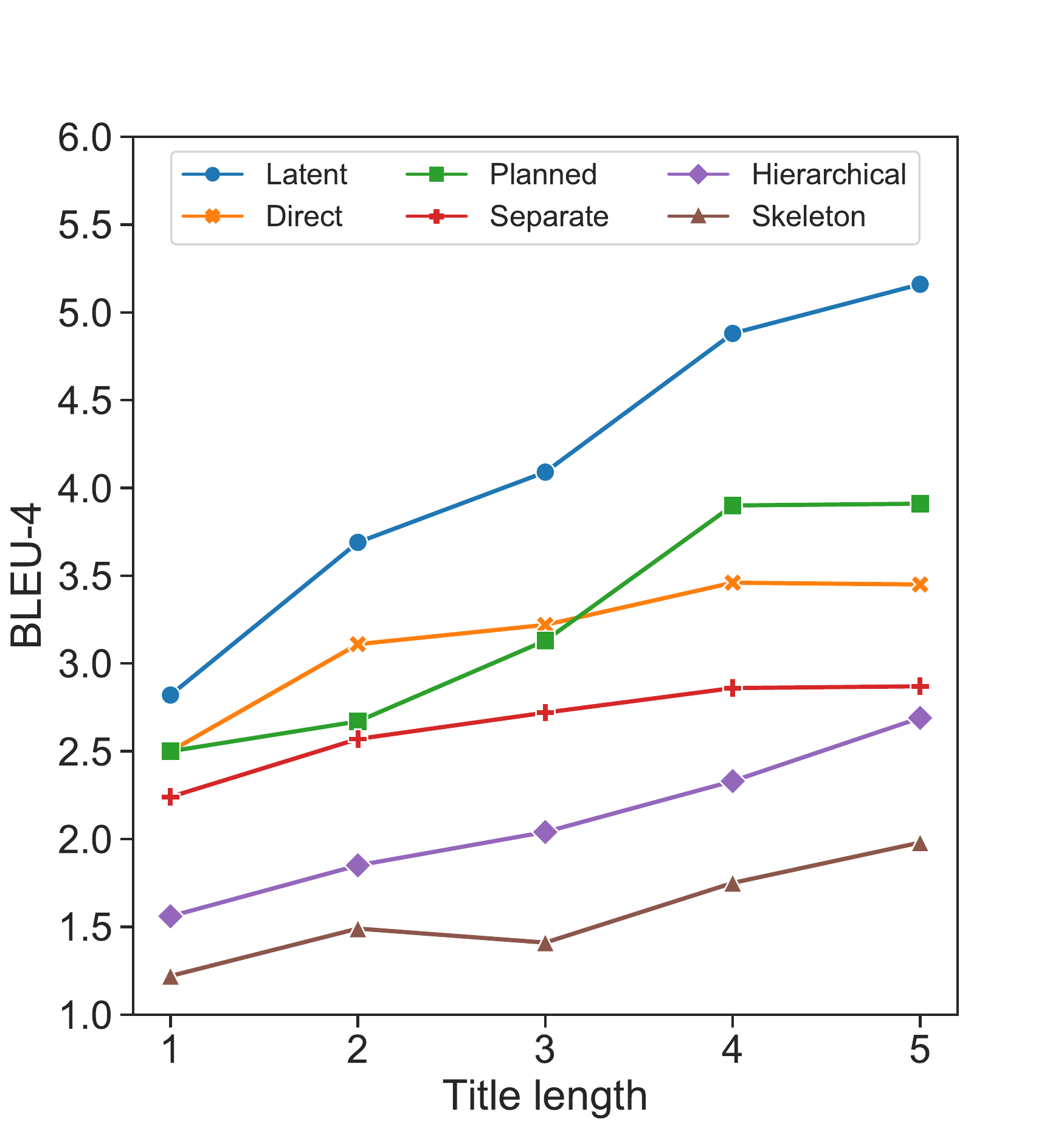}
    \caption{Effect of title length.}
    \label{fig:length_experiment}
\end{figure}

For each test set, we randomly selected 100 titles and obtained 600 stories from all the six methods (i.e., five baselines and our method). Then, we asked annotators to score the stories in terms of fluency, relevance, and coherence. Note that all the annotators have linguistic background. The stories from six methods were shuffled. The correspondences between methods and stories were invisible to the annotators. Each story was scored by three annotators. The scores were averaged to give the final score. We also computed the geometric mean of the above three metrics to measure the overall performance.

\begin{table*}[!t]
    \centering
    \scalebox{1}{
    \begin{tabular}{|l||cccc|cccc|}
    \hline
        \multirow{2}{*}{Model} & \multicolumn{4}{c|}{ROCStories} & \multicolumn{4}{c|}{VIST} \\
        \cline{2-9}
        & {F $\uparrow$} & {R $\uparrow$} & {C $\uparrow$} & {M $\uparrow$} &{F $\uparrow$} & {R $\uparrow$} & {C $\uparrow$} & {M $\uparrow$} \\
    \hline \hline
    \textproc{Direct} & 6.47 & 5.06 & 5.98 & 5.86 & 5.20 & 4.67 & 5.32 & 5.07 \\
    \textproc{Skeleton} & 6.00 & 3.93 & 4.06 & 4.76 & 5.22 & 4.47 & 5.01 & 4.91 \\
    \textproc{Hierarchical} & 5.69 & 4.26 & 4.44 & 4.84 & 5.42 & 4.85 & 5.39 & 5.22 \\
    \textproc{Separate} & 6.47 & 4.35 & \textbf{6.15} & 5.73 & 5.93 & 4.56 & 5.75 & 5.44 \\
    \textproc{Planned} & 6.35 & 4.43 & 5.37 & 5.44 & 5.86 & 4.94 & 5.52 & 5.45 \\
    \hline
    \textproc{Latent} & \textbf{6.78} & \textbf{5.80} & 6.13 & \textbf{6.25} & \textbf{6.43} & \textbf{5.69} & \textbf{6.35} & \textbf{6.17} \\
    \hline
    \end{tabular}
    }
    \caption{Human evaluation on the ROCStories and VIST datasets. We use ``F'' to denote fluency, ``R'' to denote relevance, ``C'' to denote coherence, and ``M'' to denote the geometric mean of fluency, relevance, and coherence.}
    \label{tab:result_manual}
\end{table*}

\subsection{Evaluation Results}

\begin{table}[!t]
\centering
\scalebox{1}{
\begin{tabular}{|l|c|c|c|}
\hline
Model & BLEU & METEOR & ROUGE-L \\
\hline  \hline
\textproc{Latent} & \textbf{3.70} & \textbf{10.61} & \textbf{23.06} \\
~~$-$\,PM & 3.01 & \,~9.95 & 22.09 \\
~~$-$\,SM & 2.90 & \,~9.83 & 21.97 \\
~~$-$\,PT & 3.39 & 10.32 & 22.49 \\
\hline
\end{tabular}
}
\caption{The results of ablation study. We use ``PM'' to denote the planning model, ``SM'' to denote the summarization model, and ``PT'' to denote pre-training.} \label{table:ablation}
\end{table}

\subsubsection*{Automatic Evaluation}
Table~\ref{tab:auto_results} lists the results of automatic evaluation on the ROCStories and VIST datasets. \textproc{Skeleton} generally performs worse than other baselines except that it obtains lower perplexities than \textproc{Separate} on both datasets. Surprisingly, although \textproc{Direct} does not use any high-level plot, it outperforms other baselines in terms of BLEU, METEOR, and perplexity. One possible reason is that the stories in both datasets are relatively short. The superiority of hierarchical models over direct mapping might increase when generating very long stories. Our approach significantly outperforms all baselines in terms of all listed metrics on both datasets ($p<0.01$) \footnote{We did the statistical significance tests by paired bootstrap resampling \cite{koehn:emnlp04}.}.

\begin{table}[!t]
\centering
\scalebox{1}{
\begin{tabular}{|l|c|c|c|}
	\hline
	Training & BLEU & METEOR & ROUGE-L \\
	\hline
	\hline
	VI & 3.41 & 10.23 & 21.47 \\
	\hline
	\textproc{Latent} & \textbf{3.70} & \textbf{10.61} & \textbf{23.06} \\
	\hline
\end{tabular}
}
\caption{Comparison with variational inference on the ROCStories dataset. We use ``VI'' to denote variational inference (see Section~\ref{fn:vi}).}
	\label{tab:train_obj}
\end{table}

\subsubsection*{Human Evaluation}
Table~\ref{tab:result_manual} shows the results of human evaluation on the ROCStories and VIST datasets. While all the six methods achieve close fluency scores because of the use of neural networks, the gaps on relevance and coherence are much larger. On the ROCStories dataset, our approach achieves significantly better results than all baselines in terms of fluency, relevance, and G-Mean except that \textproc{Separate} obtains a slightly higher coherence score than our approach. On the VIST dataset, our method achieves the highest scores on all the metrics by large margins. These differences are statistically significant ($p<0.01$).

\subsubsection*{Ablation Study}

Table~\ref{table:ablation} shows the results of ablation study. Our approach reduces to \textproc{Direct} by removing the planning model. Note that the summarization model is also disabled in this case. Only discarding the summarization model leads to the worst results. This suggests that the summarization model is critical for improving the performance of our approach: it is difficult to learn a good planning model without the guidance of the summarization model.  Besides, the performance of our method also drops without pre-training, which indicates the effectiveness of pre-training in our approach.

\subsubsection*{Comparison with Variational Inference}
Table~\ref{tab:train_obj} shows the comparison between our approach and variational inference on the ROCStories dataset. We find that training by variation inference deteriorates the performance of our model. One possible reason is that in variational inference the generation model is not exposed to the outlines generated by the planning model during training. During inference, however, the generation model takes the outline output by the planning model as one input. This mismatch between training and inference seems to be harmful for the NSG task. In contrast, the training objective in Eq.~(\ref{eq:train_obj}) results in significantly better performance in terms of all metrics ($p<0.01$).

\subsubsection*{Effect of Title Length}

Figure~\ref{fig:length_experiment} shows the effect of title length on the ROCStories dataset. In the test set, there are 2,546 stories for 1-word titles, 4,110 stories for 2-word titles, 1,989 stories for 3-word titles, 808 stories for 4-word titles, and 252 stories for 5-word titles. The average BLEU score was calculated for each length. We find that all methods achieve higher average BLEU scores for longer titles. This is because the information gap between title and story can be reduced if the title contains more information. Our approach outperforms all baselines over all lengths and the gaps seem to be enlarged on longer titles.

\begin{figure}[t]
    \centering
    \includegraphics[width=0.48\textwidth]{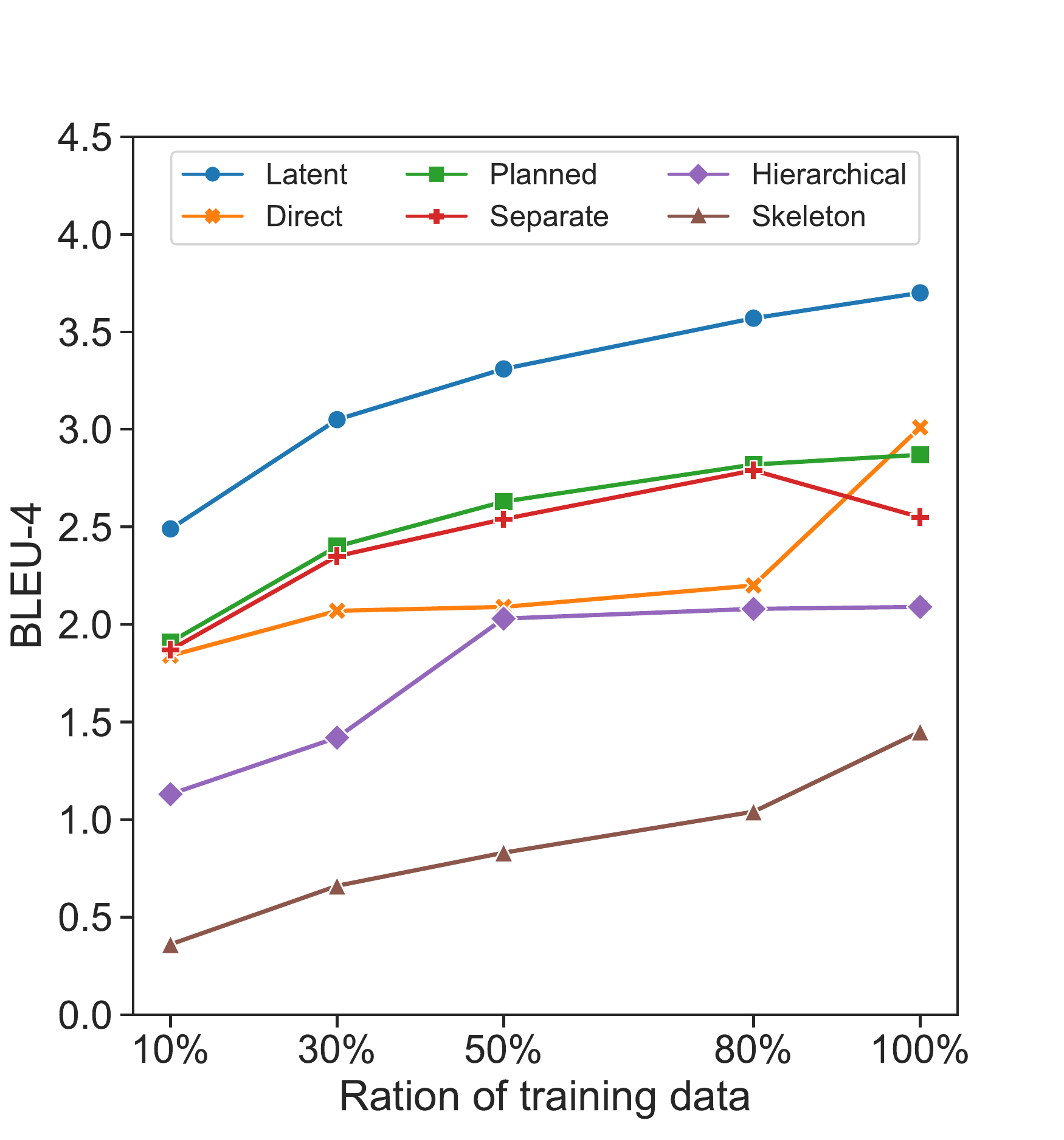}
    \caption{Effect of training corpus size.}
    \label{fig:train_size}
\end{figure}

\begin{table*}[!t]
\centering
\begin{tabular}{|l|c|}
\hline
Model & Story \\
\hline
\hline
\textproc{Direct} & \multicolumn{1}{m{11cm}|}{Tom was at the bar. He was drinking a bit too much. He got a bit too drunk. He got a bit too drunk and \textbf{passed out}. He had to be taken back home.} \\
\hline
\textproc{Skeleton} & \multicolumn{1}{m{11cm}|}{Patty was at a college. She was looking for a new camera. Patty saw a woman. She was looking for a ring.} \\
\hline
\textproc{Hierarchical} & \multicolumn{1}{m{11cm}|}{I was on a road trip with my family. We had been looking for hours. The train was on the side of the road. The driver had to come to the step to get it. The driver did not want to move back in that car.} \\
\hline
\textproc{Separate} & \multicolumn{1}{m{11cm}|}{Maria was at the grocery \textbf{store}. She was shopping for a new brand. Suddenly, she realized that she forgot her wallet. She looked everywhere for it. Luckily, she was able to find it in her pocket.} \\
\hline
\textproc{Planned} & \multicolumn{1}{m{11cm}|}{I went to the \textbf{store} to buy a new bag. I wanted to buy a new one. I went to the \textbf{store} and bought a new one. I went to the \textbf{store} to buy a new one. I bought it and went to the \textbf{store}.} \\
\hline
\textproc{Latent} & \multicolumn{1}{m{11cm}|}{Patty was at the grocery \textbf{store}. She was carrying a glass of milk. Patty \textbf{fell to} \textbf{the ground}. The ice cream fell on the ground. Patty had to go to the hospital to get stitches.} \\
\hline	
\end{tabular}
\caption{Stories generated by all methods given a title ``Fainting in the store''. Words that are closely related to the title are highlighted in bold.} \label{tab:label_case_study}
\end{table*}

\begin{table*}[!t]
\centering
\begin{tabular}{|l|l|l|}
\hline
Title & \multicolumn{2}{c|}{Back to School} \\
\hline 
\multirow{3}{*}{Story} & \multicolumn{2}{l|}{Cabot's summer break was finally over. She had a wonderful summer this year. She} \\
 & \multicolumn{2}{l|}{can't wait to see her friends at school. It is hard to get to sleep the night before classes} \\
 & \multicolumn{2}{l|}{start. After a good breakfast, she was finally an official eighth grader.} \\
\hline
\multirow{4}{*}{Outlines} & \multirow{2}{2.2cm}{Iteration 0} & (S) Cabot's summer break was finally over. \\
 & & (P)  The kids went to school.\\
 \cline{2-3}
 & \multirow{2}{2.2cm}{Iteration 32K} & (S) Cabot's summer break was finally over. \\
 & & (P) She was excited to go back to school after the summer holidays. \\
\hline
\end{tabular}
\caption{The outlines learned by the summarization and planning models during training. We use ``(S)'' and ``(P)'' to denote the outlines generated by the summarization and planning models, respectively.} \label{tab:analysis_outline}
\end{table*}

\subsubsection*{Effect of Training Corpus Size}

Figure~\ref{fig:train_size} shows the effect of training corpus size. To investigate how the performance is influenced by the size of training data, we randomly extracted part of the data from the original training set of the ROCStories dataset in the proportion of 10\%, 30\%, 50\%, and 80\%, respectively. It is clear that using more training data is beneficial for improving BLEU scores. We find that our approach keeps the superiority over all baselines consistently over different sizes of training data.

\subsubsection*{Comparison of Generated Stories}

Table~\ref{tab:label_case_study} shows the stories generated by all methods given the title ``Fainting in the store''. The story generated by \textproc{Direct} includes a phrase ``passed out'' that is similar to ``fainting'' but ignores ``store''. The story generated by \textproc{Skeleton} is fluent and coherent but is totally unrelated to the title because both ``fainting'' and ``store'' do not appear in the story. Despite containing the keyword ``store'' that is related to the title, the story generated by \textproc{Planned} lacks the relevance to ``fainting'' and suffers from sentence repetition.
%The story generated by \textproc{Planned} has the keyword ``store'' which is related to the title, but lacks relevance to ``fainting'' and also suffers from sentence repetition.

The story generated by \textproc{Hierarchical} %suffers from the same problem as 
shares the same problem with that of \textproc{Skeleton} because it is neither related to ``fainting'' nor related to ``store''. Using outlines to provide a high-level plot, \textproc{Separate} generated a fluent and coherent story. However, there is no ``fainting'' in the story. One possible reason is that \textproc{Separate} uses an off-the-shelf text summarizer to extract an outline only from the story. As the stories in the two datasets are relatively short, direct mapping between titles and stories achieves much better performance than other baselines. In addition, \textproc{Direct} avoids the error propagation problem (i.e., noisy plots inevitably leads to degenerate stories) by direct mapping.

By learning to generate outlines guided by the summarizarion model, our approach is able to generate a fluent story centered on the title. A problem with the story is that ``ice cream'' seems to be irrelevant and should be replaced with ``milk''. How to model such cross-sentence dependencies in stories is another key problem in story generation. In this work, such dependencies are modeled using self-attention \cite{Vaswani:17}. It is necessary to develop more powerful mechanisms to further improve the coherence at the story level.

\subsubsection*{Learned Outlines during Training}

Table~\ref{tab:analysis_outline} shows  the outlines learned by the summarization and planning models at different iterations during training. We find the outlines learned by the summarization model often remain identical during training. On the contrary, the outlines generated by the planning model keep changing during training. While the predicted outline is ``The kids went to school.'' at iteration 0 in the beginning, the prediction is changed to ``She was excited to go back to school after the summer holidays''. According to Eq.~(\ref{eq:train_obj}), the planning model is involved in both terms of the training objective, which suggests that the outline generated by the planning model should not only be close to that of the summarization model, but also enable the generation model to generate the ground-truth story. We find that the outline generated by the planning model at iteration 32K contains more useful information about the story than that of the summarization model. Besides filling the information gap between titles and stories, these predicted outlines are interpretable to humans and help them better understand how NSG model works.
%As the generated outlines can be easily inspected, our method has a certain degree of interpretability.

\section{Conclusion and Future Work}
We have proposed a method for learning to predict explainable plots for neural story generation for generating fluent, coherent, and reasonable stories. Different from existing methods, our model is capable of learning to automatically produce an explainable high-level plot to bridge the information gap between the title and story. Experiments on two benchmark datasets show that our method outperforms state-of-the-art approaches to neural story generation in both automatic and human evaluations.

In this work, we have focused on using a one-sentence outline to bridge the information gap between a title and a short story. How to model the cross-sentence dependencies in a long story still remains a severe challenge. It is important to develop more effective mechanisms to improve story-level coherence. We plan to extend our method to multiple outline generation for generating very long stories in the future.

\bibliography{reference}
\bibliographystyle{acl_natbib}
\end{document}